\documentclass[runningheads]{llncs}
\usepackage[T1]{fontenc}
\usepackage{graphicx}
\usepackage{subfig}
\usepackage{multirow}
\usepackage{amsmath}
\usepackage{amssymb}
\usepackage{wrapfig}
\usepackage{booktabs}
\usepackage{enumitem}
\usepackage{algorithm}
\usepackage{algorithmic}
\usepackage{newfloat}
\usepackage{listings}
\usepackage{xcolor}
\usepackage{bbding}

\begin{document}
\title{NK-GAD: Neighbor Knowledge-Enhanced Unsupervised Graph Anomaly Detection}
\titlerunning{NK-GAD: Unsupervised Graph Anomaly Detection}
% \author{Anonymous}
% \institute{}
\author{Zehao Wang\inst{1,2}
% \orcidID{0000-0003-4267-5359}
\and
Lanjun Wang\inst{2}(\Envelope)
% \orcidID{0000-0002-7696-5330}
}
\authorrunning{Zehao Wang and Lanjun Wang}
\institute{College of Intelligence and Computing, Tianjin University, Tianjin, China \and
School of New Media and Communication, Tianjin University, Tianjin, China
\email{wzhrslh@gmail.com, wanglanjun@tju.edu.cn}}
\maketitle              % typeset the header of the contribution
\begin{abstract}
Graph anomaly detection aims to identify irregular patterns in graph-structured data. Most unsupervised GNN-based methods rely on the homophily assumption that connected nodes share similar attributes. However, real-world graphs often exhibit attribute-level heterophily, where connected nodes have dissimilar attributes. Our analysis of attribute-level heterophily graphs reveals two phenomena indicating that current approaches are not practical for unsupervised graph anomaly detection: 1) attribute similarities between connected nodes show nearly identical distributions across different connected node pair types, and 2) anomalies cause consistent variation trends between the graph with and without anomalous edges in the low- and high-frequency components of the spectral energy distributions, while the mid-part exhibits more erratic variations. Based on these observations, we propose NK-GAD, a neighbor knowledge-enhanced unsupervised graph anomaly detection framework. NK-GAD integrates a joint encoder capturing both similar and dissimilar neighbor features, a neighbor reconstruction module modeling normal distributions, a center aggregation module refining node features, and dual decoders for reconstructing attributes and structures. Experiments on seven datasets show NK-GAD achieves an average 3.29\% AUC improvement.

\keywords{Unsupervised Graph Anomaly Detection  \and Graph Mining \and Unsupervised Node Classification.}
\end{abstract}
\section{Introduction}
Graph anomaly detection is crucial for identifying irregular or suspicious patterns in graph-structured data, with applications in fields like social networks~\cite{zardi2023anomaly} and financial transactions~\cite{thilagavathi2024ai}.
Existing works~\cite{liu2022bond} typically categorize anomalies in graph-structured data into two types: 1) contextual anomalies, where node attributes deviate from the normal nodes, and 2) structural anomalies, where irregularities occur in the edges between nodes. 
However, due to the high cost of obtaining labeled graph data in real-world applications, recent advancements~\cite{he2024ada,roy2024gad} have increasingly focused on unsupervised detection methods.

% Dilemma of GAD
Recently, the unsupervised graph anomaly detection methods~\cite{qiao2024deep} are mainly built on graph neural networks (GNNs)
~\cite{wu2020comprehensive}, which rely on a message-passing mechanism to aggregate information from neighboring nodes and capture the underlying patterns of normal behavior.
Prior research has demonstrated that the performance of Graph Neural Networks (GNNs) is largely governed by the homophily assumption~\cite{luan2024graph}, where connected nodes with the same labels or similar features. 
However, recent studies~\cite{he2024ada} reveal the existence of label-level heterophily, where connected nodes belong to different classes (e.g., anomalous vs. normal). 
This structural characteristic severely undermines the effectiveness of GNN-based models, which rely on neighborhood similarity for message propagation. 
Motivated by the observed right-shifting effect~\cite{gao2023addressing,tang2022rethinking}, where low-frequency energy gradually shifts toward the high-frequency spectrum as the anomaly degree increases, subsequent works~\cite{he2024ada,tang2022rethinking} have sought to alleviate the negative influence of anomalous nodes on normal ones. These approaches typically modify graph structures or node representations to reduce harmful information transfer between the two node types.

\begin{figure}[t!]
	\centering
    \subfloat[Weibo]{
		\includegraphics[width=0.4\textwidth]{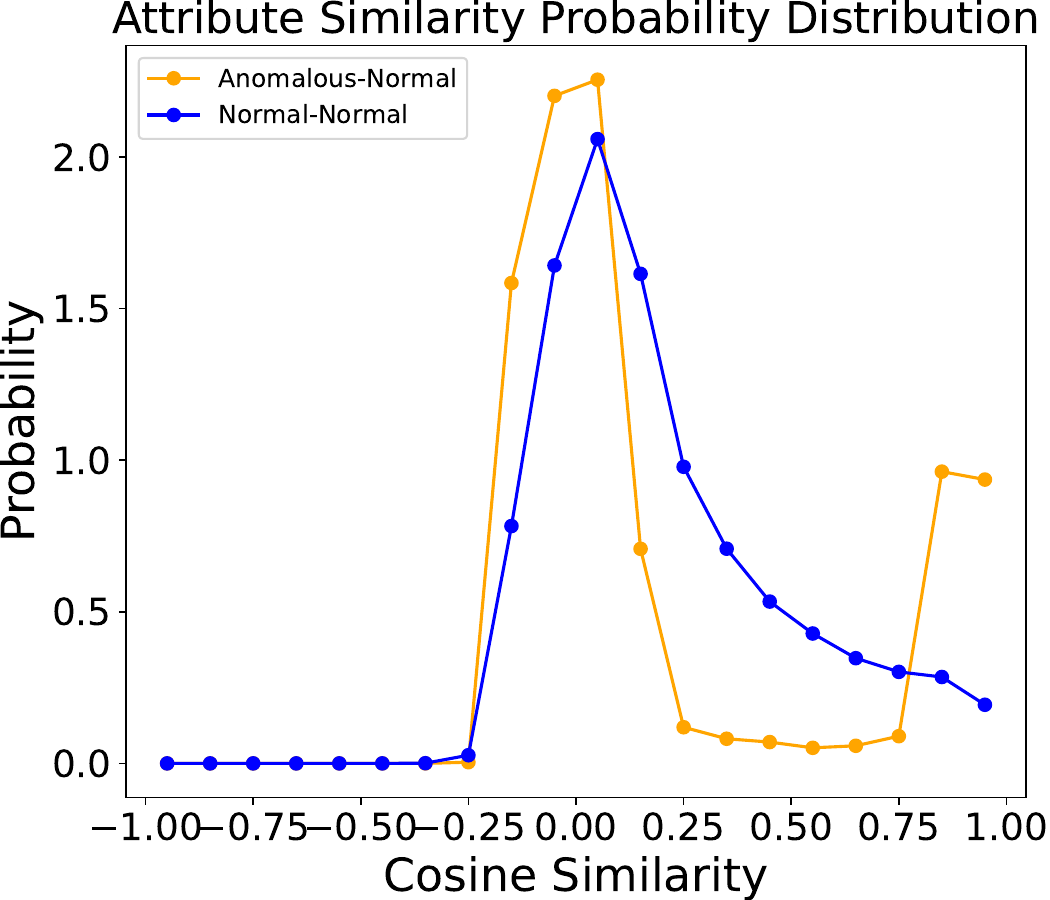}
		\label{dis_weibo}}
        \hspace{-2mm}
	\subfloat[Books]{
		\includegraphics[width=0.4\textwidth]{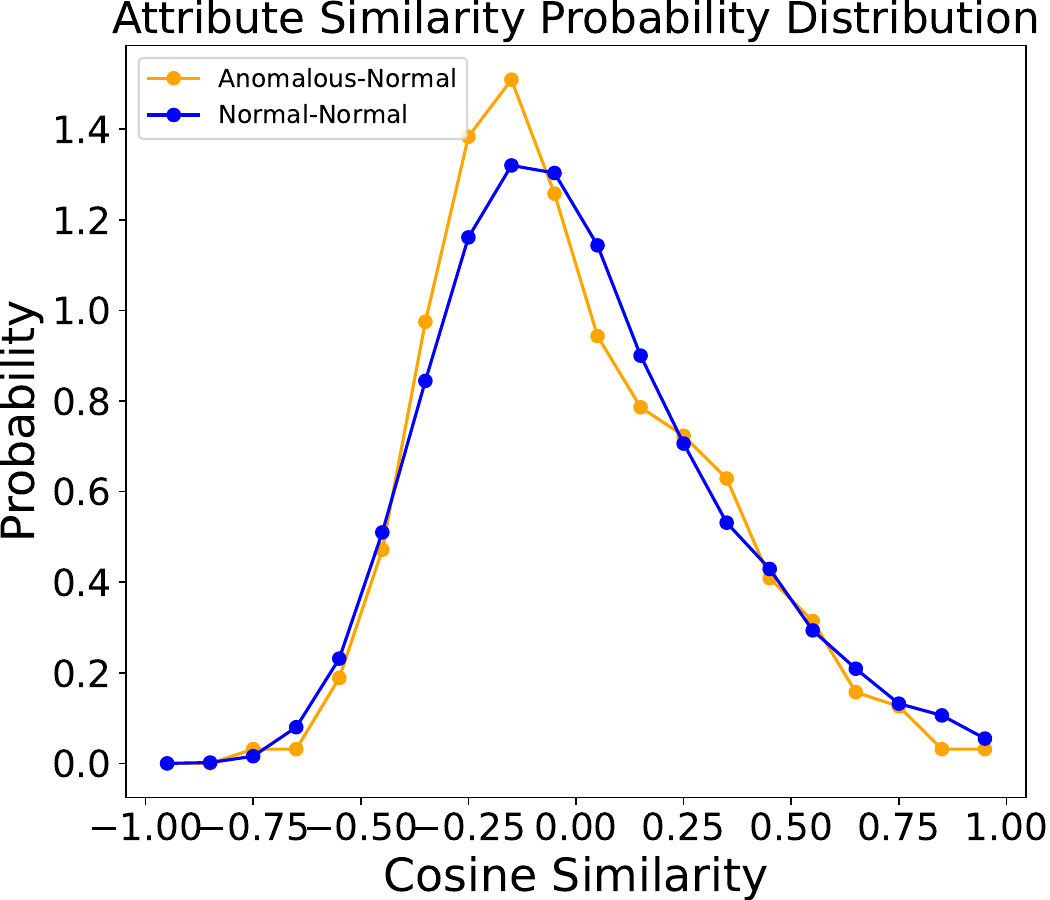}
		\label{dis_books}} 
	\caption{
    Attribute similarity distributions of anomalous–normal and normal–normal node pairs in the Weibo and Books datasets mostly fall within the low cosine similarity range, indicating strong feature dissimilarity between connected nodes. Moreover, the two distributions largely overlap across pair types.
    }
    \label{fig:feature_dis}
\end{figure}

\begin{figure}[t!]
    \centering
    \subfloat[Weibo]{
		\includegraphics[width=0.42\textwidth]{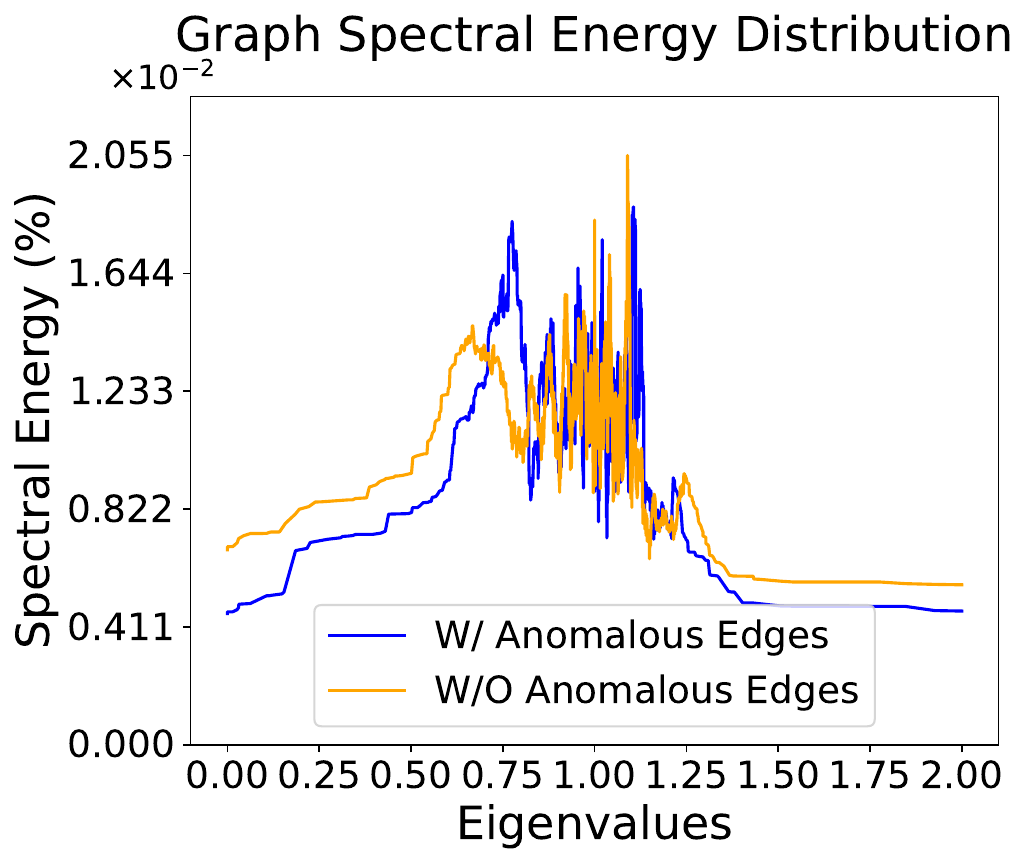}
		\label{spectrum_weibo}} \hspace{-2mm}
	\subfloat[Books]{
		\includegraphics[width=0.4\textwidth]{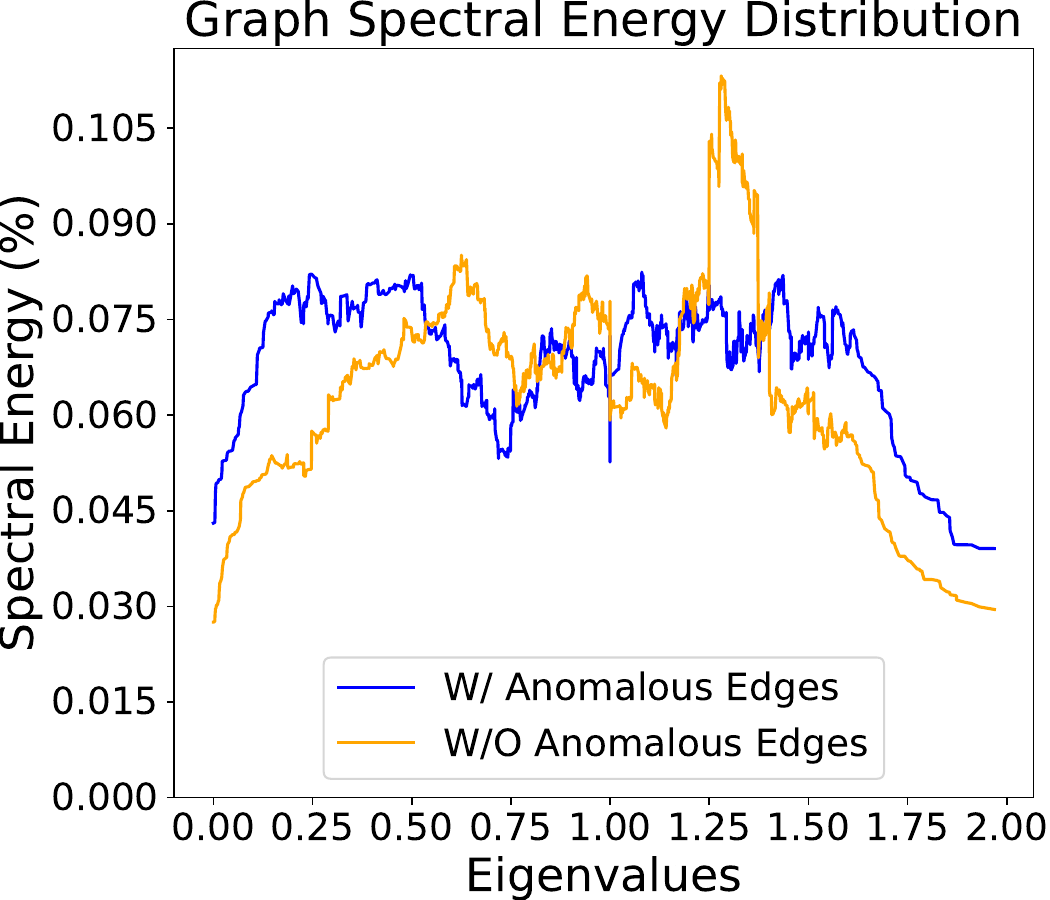}
		\label{spectrum_reddit}} 
	
\caption{
    Graph spectral energy distribution for the graphs w/ and w/o anomalous edges in the Weibo and Books datasets.
    Compared to the graphs w/o anomalous edges, which remove the edges between normal and anomalous nodes~\cite{he2024ada},
    the spectral energy distributions of the graphs w/ anomalous edges show consistent variation trends in low- and high-frequency components, while the mid-frequency components exhibit more erratic variations.
    }
    \label{fig:spectral_energy}
\end{figure}

However, the existing unsupervised methods~\cite{he2024ada,roy2024gad} overlook the attribute-level heterophily~\cite{pmlr-v235-wang24u,luan2024graph} present in real-world graph-structured data. In detail, the attribute-level heterophily refers to the dissimilarity between connected nodes in the spatial domain, where the majority of the cosine similarity values fall within the low range of [0, 0.25] (Fig.~\ref{fig:feature_dis}). 
It is also manifested in the spectral domain of the graphs, where spectral energy is distributed across the mid-frequency components, such as over 90\% of the spectral energy in the Weibo dataset is concentrated in the range with eigenvalues between [0.75, 1.25) (Fig.~\ref{fig:spectral_energy}).

Furthermore, based on Figs.~\ref{fig:feature_dis} and \ref{fig:spectral_energy},  we recognize two phenomena that indicate the existing methods are not practical for unsupervised graph anomaly detection.
% spatial domain
Firstly, in the spatial domain (Fig.~\ref{fig:feature_dis}), the attribute similarities of the connected nodes across different connected node pair types (anomalous-normal and normal-normal) display nearly identical distributions.
This observation indicates that it is not reasonable to judge whether the edges or attributes are anomalous solely based on the similarity of the connected nodes' attributes.
However, existing unsupervised methods~\cite{he2024ada,tang2022rethinking} modify edges and attributes between connected nodes with highly dissimilar attributes only based on such similarity. 
Thus, these methods remove anomalous edges and attributes as well as normal ones in attribute-level heterophily graphs, resulting in the loss of information related to the normal patterns between neighbors.

Secondly, in the spectral domain (Fig.~\ref{fig:spectral_energy}), anomalies in attribute-level heterophily graphs lead to consistent variation trends in the low- and high-frequency components of the spectral energy distributions~\cite{tang2022rethinking}, while the mid-frequency components exhibit more erratic variations.
For instance, the Weibo datasets~\cite{liu2022bond} show increases in spectral energy in the low- and high-frequency ranges after removing abnormal edges.
This observation implies that the distinction between normal and anomalous patterns is clearer in the low- and high-frequency components than in the mid-frequency range. 
However, existing methods~\cite{he2024ada,tang2022rethinking} neglect the importance of the high-frequency component.
For instance, Tang et al.~\cite{tang2022rethinking} focus on processing the mid-frequency component to extract both the normal and anomalous patterns, while He et al.~\cite{he2024ada} aim to mitigate the impact of anomalous nodes by reducing high-frequency energy.
Thus, they do not consider the dissimilar attributes of neighbors, and it is challenging to distinguish between normal and anomalous patterns in unsupervised settings.

% model design
As a result, in this paper, we focus on incorporating neighbor knowledge, e.g. neighbor features and the corresponding distributions, to improve the ability to detect graph anomalies in an unsupervised setting.
We design a novel framework, named \textbf{N}eighbor \textbf{K}nowledge-enhanced unsupervised \textbf{G}raph \textbf{A}nomaly \textbf{D}etection (NK-GAD), which consists of a joint graph convolutional encoder, a neighbor reconstruction module, a center aggregation module, and two decoders. 
The joint graph convolutional encoder consists of a low-pass filter and a high-pass filter to parallel extract patterns of both the similar and dissimilar attributes of neighbors. 
To deal with two types of anomalies, we design two modules for structural and contextual anomalies, respectively.
In detail, the neighbor reconstruction module enhances node representations by reconstructing neighboring feature distributions based on the hidden representation in the encoder, aiding in the detection and removal of structural anomalies.
The center aggregation module refines node features by leveraging reconstructed neighbor distributions, filtering out contextual anomalies. 
Finally, two decoders reconstruct node attributes and the adjacency matrix to compute anomaly scores and total loss.
The contributions are summarized as follows:
\begin{itemize}[nosep]
    \item We present two observations from real-world datasets, analyzed from both spatial and spectral perspectives, that highlight the limitations of existing methods in graph anomaly detection. Building on these insights, we propose a novel framework, NK-GAD, which leverages neighbor knowledge to detect the graph anomaly.
    \item We introduce a joint graph convolutional encoder that captures both similar and dissimilar neighbor features for extracting normal patterns across diverse attributes.
    \item To mitigate the impact of different kinds of anomalies on the normal nodes, NK-GAD incorporates a neighbor reconstruction module to extract normal patterns and a center aggregation module to refine the central node's features using reconstructed neighbor distributions.
    \item Extensive experiments on seven real-world datasets show the effectiveness of NK-GAD, with up to a 3.29\% improvement in AUC over SOTA methods.
\end{itemize}

\section{Related Work}

\subsection{Graph Neural Networks}

Graph Neural Networks (GNNs) have garnered significant attention for their ability to model graph-structured data effectively across various domains, such as traffic prediction~\cite{shaygan2022traffic} and semantic search~\cite{zhang2021alicg}. GNNs exploit the graph structure by aggregating information from neighboring nodes, enabling them to capture local patterns and dependencies efficiently.
A key principle is the homophily assumption~\cite{luan2024graph}, where connected nodes share similar attributes or labels.
Models such as GCNs~\cite{kipf2016semi} and GATs~\cite{velivckovic2017graph} leverage message passing to propagate information along edges, reinforcing node representations for tasks like node classification~\cite{xiao2022graph} and link prediction~\cite{kumar2020link}.
However, the homophily assumption limits the performance of GNNs on attribute-level heterophily graphs~\cite{pmlr-v235-wang24u,luan2024graph}, where connected nodes often have dissimilar attributes or belong to different labels.
Recent studies~\cite{luan2024graph,pmlr-v235-wang24u} show that attribute-level heterophily is not inherently detrimental, as it can provide valuable insights for uncovering hidden patterns in real-world graph-structured data.

Inspired by the findings, we analyze real-world datasets in the graph anomaly detection task and observe two phenomena (Fig.~\ref{fig:feature_dis} and Fig.~\ref{fig:spectral_energy}) that motivate us to combine neighbor knowledge to improve the ability to detect anomalies.

\subsection{Unsupervised Graph Anomaly Detection}
Graph anomaly detection seeks to identify anomalies in graph-structured data~\cite{liu2022bond}, a crucial task with applications in fraud detection~\cite{liu2021pick} and social network analysis~\cite{hc2023retracted}.
Due to the rarity of anomalies and the high cost of obtaining labeled graph data in real-world applications, recent research has increasingly focused on unsupervised detection methods.
Traditional algorithms~\cite{li2017radar,liu2022bond} show good performance on certain datasets that own obvious differences between normal and anomalous nodes.
Compared to these traditional algorithms, deep learning-based detection models often exhibit better generalization~\cite{liu2022bond}.
With the deepening research into graph-structured data, the emergence of Graph Neural Networks (GNNs)~\cite{kipf2016semi} has revolutionized graph anomaly detection. 
GNNs-based graph anomaly detection methods enable the learning of rich, task-specific representations of nodes and edges ~\cite{he2024ada,roy2024gad}. 
Works like ADA-GAD of He. et al. and AnomalyADE of Fan. et al. utilize GNN-based architectures to detect anomalies by reconstructing node attributes and adjacency matrices, identifying discrepancies between reconstructed and observed data. GAD-NR~\cite{roy2024gad} and SmoothGNN~\cite{dong2025smoothgnn} focus on leveraging neighbor feature information to detect anomalies.

However, existing studies on unsupervised graph anomaly detection overlook the attribute-level heterophily of real-world graph-structured datasets. Moreover, our observations show that these methods are not suitable for attribute-level heterophily graphs. Thus, we propose NK-GAD, a framework designed for unsupervised graph anomaly detection on attribute-level heterophily graphs.

\section{Problem Formulation}
A graph with node attributes is formally defined as $\mathcal{G} = (\mathcal{V}, \mathcal{E}, \mathbf{X})$, where $\mathcal{V} = \{v_1, v_2, \dots, v_{|\mathcal{V}|}\}$ is the set of nodes, $\mathcal{E} = \{e_1, e_2, \dots, e_{|\mathcal{E}|}\}$ is the set of edges, and $\mathbf{X} \in \mathbb{R}^{|\mathcal{V}| \times dim}$ is the node attribute matrix. Specifically, each edge $e \in \mathcal{E}$ is represented as a tuple $(v_i, v_j)$, where $v_i, v_j \in \mathcal{V}$ are the nodes connected by the edge. The structural relationships captured in $\mathcal{E}$ can alternatively be expressed as a binary adjacency matrix $\mathbf{A} \in \mathbb{R}^{|\mathcal{V}| \times |\mathcal{V}|}$, where $\mathbf{A}_{i, j} = 1$ indicates the presence of an edge $(v_i, v_j) \in \mathcal{E}$, and $\mathbf{A}_{i, j} = 0$ otherwise. In the attribute matrix $\mathbf{X}$, the $i$-th column is denoted as $\mathbf{x}_{i} \in \mathbb{R}^{dim}$ which represents the attribute vector of the corresponding node $v_i$. 
The degree matrix is defined as $\mathbf{D} = \mathrm{diag}(d_1, \dots, d_{|\mathcal{V}|})$, where $d_i = \sum_{j=1}^{|\mathcal{V}|} \mathbf{A}_{i,j}$ is the degree of node $v_i$. 
The symmetric normalized Laplacian matrix of a graph $\mathcal{G}$ is defined as $\mathbf{L} = \mathbf{D}^{-1/2} (\mathbf{D} - \mathbf{A}) \mathbf{D}^{-1/2}$. Since $\mathbf{L}$ is positive semi-definite and symmetric, it can be decomposed as $\mathbf{L} = \mathbf{U} \Lambda \mathbf{U}^\top$, where $\Lambda = \{\lambda_{1}, \lambda_{2}, \dots, \lambda_{|\mathcal{V}|}\}$ are the eigenvalues, and each column $\mathbf{u}_{i}$ is the unit eigenvector corresponding to eigenvalue $\lambda_{i}$. 
Additionally, the neighbors of node $v_i$ are denoted by $\mathcal{Z}(i) = \{v_j \mid (v_i, v_j) \in \mathcal{E}\}$. 
Referring to~\cite{roy2024gad,tekin2023crime}, we regard the neighbor feature distribution of the node $v_i\in\mathcal{V}$ as a multi-variate Gaussian distribution $\mathbb{P}_{\mathcal{Z}(i)}\sim \mathcal{N}(\mathbf{\mu}_i, \mathbf{\Sigma}_i)$, where $\mathbf{\mu}_i\in\mathbb{R}^{dim}$ is the vector of mean of each dimension and $\mathbf{\Sigma}_i\in\mathbb{R}^{dim\times dim}$ is the covariance matrix.

In this paper, we adapt the unsupervised setting, wherein the model does not have access to node labels during training, following prior studies~\cite{he2024ada,roy2024gad}.
Given a graph $\mathcal{G} = (\mathcal{V}, \mathcal{E}, \mathbf{X})$, the goal of a reconstruction-based graph anomaly detection model $f$ is to assign an anomaly score $s_i$ to each node $v_i \in \mathcal{V}$ based on the input graph $\mathcal{G}$. Formally, this can be expressed as:
\begin{equation}
f: \mathcal{G} \rightarrow \{s_1, s_2, \dots, s_i, \dots, s_{|\mathcal{V}|} \}, \quad s_i \in \mathbb{R}_{\geq 0}
\end{equation}
where the anomaly score $s_i$ quantifies the degree of abnormality of the node $v_i$. A higher score indicates a higher likelihood of the node being anomalous, whereas a lower score suggests it is more likely to be normal.

\section{Methodology}\label{sec:method}

\begin{figure*}[ht!]
    \centering
    \includegraphics[width = 0.99\textwidth]{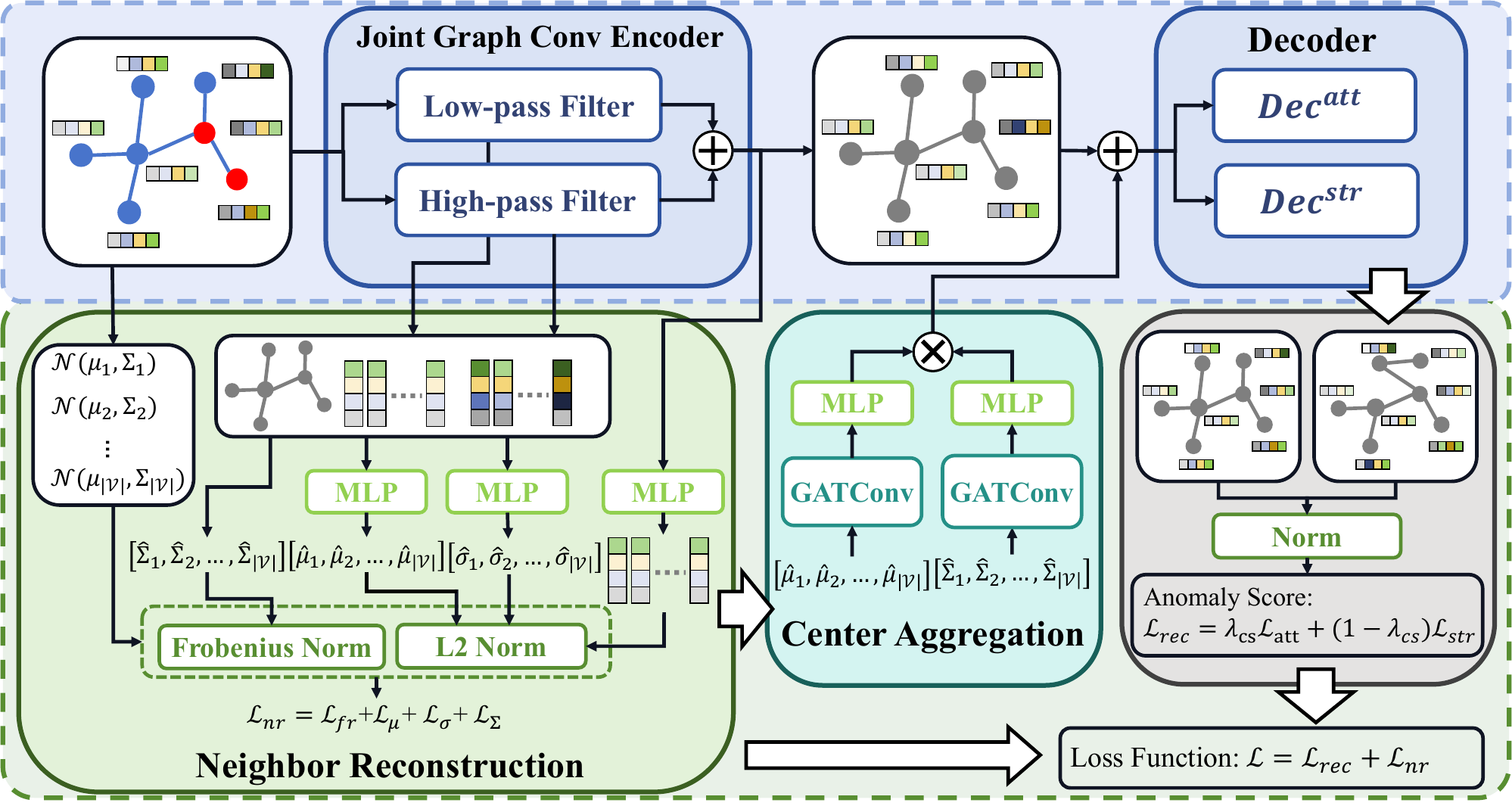}
    \caption{Overview of NK-GAD. NK-GAD integrates four core components to reconstruct graph data, subsequently computing anomaly scores and total loss for detection.
    Red, blue, and gray nodes highlight anomalies, normal instances, and unseen labels, respectively.
    }
    \label{fig:NK-GAD}
\end{figure*}

In this section, we introduce the proposed NK-GAD, and its overview is shown in Fig.~\ref{fig:NK-GAD}. NK-GAD consists of a joint graph convolutional encoder, a neighbor reconstruction module, a center aggregation module, and two decoders. The joint graph convolutional encoder combines the low- and high-frequency components to extract patterns from similar and dissimilar attributes between central and neighboring nodes. 
Next, the neighbor reconstruction module utilizes the extracted patterns to reconstruct the feature distribution of neighbor attributes.
Minimizing the loss of the neighbor reconstruction module improves the extraction of normal patterns. Then, the center aggregation module uses these patterns to refine central node attributes. 
Finally, two decoders reconstruct node attributes and edges, and calculate the anomaly scores.

\subsection{Joint Graph Convolutional Encoder}\label{sec:joint}

Observations from the spatial domain indicate that distinguishing between normal and anomalous nodes requires further extraction of patterns contained within the features of attribute-level heterophily graphs. Meanwhile, observations from the spectral domain highlight the advantage of leveraging low- and high-frequency components to differentiate between normal and abnormal patterns in an unsupervised setting, where the low- and high-frequency components can be used to represent the similar and dissimilar attributes between neighbors, respectively. 

Motivated by these insights, to extract patterns from the features of neighbors, including both the similar and dissimilar attributes, we introduce a joint graph convolutional encoder that integrates a low-pass filter and a high-pass filter to process both frequency components in parallel.

We first project the node attributes into a hidden representation space using a linear layer, formulated as $\mathbf{H}^{0} = \text{Linear}(\mathbf{X})$, where $\mathbf{H}^{0} \in \mathbb{R}^{|\mathcal{V}| \times d}$ represents the hidden representation, and $d$ denotes the dimensionality of the hidden space.

The low-pass and high-pass filters in the joint graph convolutional encoder are based on the normalized Laplacian matrix $\mathbf{L}$, facilitating the separation of smooth and sharp transitions in the graph $\mathcal{G}$. 
Based on the Rayleigh quotient~\cite{stankovic2020vertex}, the eigenvalues of $\mathbf{L}$ are bounded in the range $[0, 2]$. These eigenvalues are used to define the graph's frequency components: low-frequency components correspond to eigenvalues in $[0, 1)$, while high-frequency components correspond to eigenvalues in $(1, 2]$.
Thus, the low- and high-pass filters are defined as follows:
\begin{equation} 
f_{low} = \mathbf{I} + \mathbf{D}^{-\frac{1}{2}} \mathbf{A} \mathbf{D}^{-\frac{1}{2}} = 2\mathbf{I} - \mathbf{L}, \quad
f_{high} = \mathbf{I} - \mathbf{D}^{-\frac{1}{2}} \mathbf{A} \mathbf{D}^{-\frac{1}{2}} = \mathbf{L}
\end{equation}

To process the hidden representations of each node, we adopt a multi-layer architecture that leverages graph convolution operations~\cite{huang2023robust,dong2020graph}. The representations are separately computed for low- and high-frequency components at each layer as follows:
\begin{align} 
\mathbf{H}_{low}^{l} &= f_{low} *_\mathcal{G} \mathbf{H}^{l-1}_{low}\mathbf{W}^{l}_{low}, \quad l = 1, 2, \dots, L  \\
\mathbf{H}^{l'}_{high} &= f_{high} *_\mathcal{G} \mathbf{H}^{l'-1}_{high}\mathbf{W}^{l'}_{high}, \quad l' = 1, 2, \dots, L'
\end{align}
where $*_\mathcal{G}$ denotes the graph convolution operator, $\mathbf{H}_{low}^{l}$ and $\mathbf{H}^{l'}_{high}$ are the representations of low- and high-frequency components at the $l$-th layer, respectively. For the first layer ($l=l'=1$), $\mathbf{H}_{low}^{0} = \mathbf{H}_{high}^{0} = \mathbf{H}^{0}$. $\mathbf{W}^{l}_{low}$ and $\mathbf{W}^{l'}_{high}$ are the learnable weight matrices for the low- and high-pass filters. Specifically, the graph convolution operations are computed in the spectral domain as:
\begin{equation} 
f_{low} *_\mathcal{G} \mathbf{H}^{l}_{low} = \mathbf{U} \left( 2\mathbf{I} - \Lambda \right) \mathbf{U}^T \mathbf{H}^{l-1}_{low}, \quad
f_{high} *_\mathcal{G} \mathbf{H}^{l'}_{high} = \mathbf{U} \Lambda \mathbf{U}^T \mathbf{H}^{l'-1}_{high}
\end{equation}

At the final layer, the representations $\mathbf{H}_{\text{low}}^{L}$ and $\mathbf{H}_{\text{high}}^{L^{\prime}}$ are combined to generate the output $\mathbf{Z}$ of the joint graph convolutional encoder as:
\begin{equation} 
\mathbf{Z} = (1 - \lambda_{joint}) \mathbf{H}_{\text{low}}^{L} + \lambda_{joint} \mathbf{H}_{\text{high}}^{L^{\prime}}
\end{equation}
where $\lambda_{joint}$ is a trade-off coefficient that balances the contributions of low- and high-frequency components.

\subsection{Neighbor Reconstruction}
\label{sec:neighbor_rec}

The patterns hidden in neighbor features can be reflected in the characteristics of neighbor feature distribution.
For a central node, similar attributes shared with neighbors represent common patterns, while dissimilar attributes capture variations. Since node feature dimensions are often interdependent~\cite{roy2024gad}, their correlations also encode shared patterns.
These patterns can be quantified via the neighbor feature distribution’s mean, standard deviation, and covariance. Features close to the mean indicate normal patterns, while large deviations reveal structural anomalies. 

This helps mitigate structural anomalies by identifying and reducing the influence of neighbors exhibiting anomalous patterns.
Therefore, we introduce a neighbor reconstruction module that leverages the characteristics of neighbor feature distributions to improve the detection of structural anomalies.

\subsubsection{Self Feature Reconstruction}

To ensure the updated representation matrix $\mathbf{Z}$ captures information about each node's neighbors, we employ a simple multi-layer perceptron (MLP) to reconstruct $\mathbf{H}^{0}$ from $\mathbf{Z}$, which can be formulated as $\hat{\mathbf{H}}^{0} = \text{MLP}(\mathbf{Z})$, where $\hat{\mathbf{H}}^{0}\in\mathbb{R}^{|\mathcal{V}|\times d}$ denotes the reconstructed representation of $\mathbf{H}^{0}$.
Then, the feature reconstruction loss $\mathcal{L}_{fr}$ is computed as $\mathcal{L}_{fr} = ||\hat{\mathbf{H}}^{0} - \mathbf{H}^{0}||_2$, where $\|\cdot\|_2$ is the L2 norm.

\subsubsection{Neighbor Feature Distribution Reconstruction}

Based on the hidden representation matrix $\mathbf{H}^{0}$, we compute the true mean, standard deviation, and covariance matrix of the neighbor feature distribution for each node in the graph.

For a given node $v_i$, the mean $\mu_i$ and standard deviation $\sigma_i$ of its neighbors' features are calculated as follows:
\begin{equation} 
    \mu_i = \frac{1}{|\mathcal{Z}(i)|} \sum_{j \in \mathcal{Z}(i)} \mathbf{h}^{0}_j, \quad
    \sigma_i = \sqrt{\frac{1}{|\mathcal{Z}(i)| - 1} \sum_{j \in \mathcal{Z}(i)} \left( \mathbf{h}^{0}_j - \mu_i \right)^2}
\end{equation}
where $\mathbf{\mu}_i, \mathbf{\sigma}_i\in\mathbb{R}^{d}$, $|\mathcal{Z}(i)|$ is the number of neighbors, and $\mathbf{h}^{0}_j$ is the feature vector of neighbor $v_j$ in $\mathbf{H}^{0}$.
Then, the covariance matrix $\mathbf{\Sigma}_i$ for the neighbor feature distribution of node $v_i$ is computed as:
\begin{equation} 
    \mathbf{\Sigma_i} = \frac{1}{|\mathcal{Z}(i)| - 1} \sum_{j \in \mathcal{Z}(i)} \left( \mathbf{h}^{0}_j - \mu_i \right) \left( \mathbf{h}^{0}_j - \mu_i \right)^T
    \label{eq:cov}
\end{equation}

Next, we predict these characteristics. 
Since the neighbor feature mean reflects smooth transitions and the standard deviation captures variations influenced by sharp transitions, we employ two MLPs to predict the mean and standard deviation for each node $v_i$, utilizing the node's low-frequency and high-frequency feature representations, $\mathbf{p}_i \in \mathbf{H}_{\text{low}}^{L}$ and $\mathbf{q}_i \in\mathbf{H}_{\text{high}}^{L^{\prime}}$, respectively:
\begin{equation} 
    \hat{\mu}_i = \text{MLP}_{\mu}\left(\mathbf{p}_i\right), \quad 
    \hat{\sigma}_i = \text{MLP}{\sigma}\left(\mathbf{q}_i\right)
\end{equation}

After calculating the mean and standard deviation for all nodes, the prediction losses for the mean and standard deviation can be formulated as follows:
\begin{equation} 
\mathcal{L}_{\mu} = \frac{1}{N} \sum_{i=1}^N \|\hat{\mu}_i - \mu_i\|_2, \quad\mathcal{L}_{\sigma} = \frac{1}{N} \sum_{i=1}^N \|\hat{\sigma}_i - \sigma_i\|_2
\end{equation}

We compute the predicted covariance matrix $\hat{\mathbf{\Sigma}}_i$ for the neighbor feature distribution of $v_i$ based on $\hat{\mathbf{\mu}_i}$ and $\hat{\mathbf{h}}^{0}_{i} \in \hat{\mathbf{H}}^{0}_{i}$ as same as Eq.~\ref{eq:cov}.

To avoid gradient explosion when calculating the KL divergence of the covariance matrix, we adopt the Frobenius norm $\|\cdot\|_{F}$ as the loss function to calculate the loss for the covariance matrix:
\begin{footnotesize}
\begin{equation} 
\mathcal{L}_{\mathbf{\Sigma}} = \frac{1}{N} \sum_{i=1}^N \|\hat{\mathbf{\Sigma}}_i - \mathbf{\Sigma}_i\|_F =  \frac{1}{N} \sum_{i=1}^N
\text{Tr}((\hat{\mathbf{\Sigma}}_i - \mathbf{\Sigma}_i)^T (\mathbf{\Sigma}_i - \hat{\mathbf{\Sigma}}_i))
\end{equation}
\end{footnotesize}

By combining the loss from self-feature reconstruction and neighbor feature distribution reconstruction, the loss for the neighbor reconstruction module is defined as follows:
\begin{equation} 
    \mathcal{L}_{nr} = \mathcal{L}_{fr} + \mathcal{L}_{\mathbf{\mu}} + \mathcal{L}_{\mathbf{\sigma}} + \mathcal{L}_{\mathbf{\Sigma}}
\end{equation}

\subsection{Center Aggregation}\label{sec:center_agg}

For each neighbor $v_j \in \mathcal{Z}(i)$ of a central node $v_i$, the node $v_i$ also serves as a neighbor of $v_j$. Thus, the normal features of $v_i$ should align with the neighbor feature distributions of all $v_j \in \mathcal{Z}(i)$. Leveraging the mean and covariance of these distributions allows smoothing and updating $v_i$’s representation, mitigating contextual anomalies.
Building on this principle, we design the center aggregation module to leverage these neighbor feature distributions and improve the central node’s feature representation.

In this module, we utilize the graph attention layer~\cite{velivckovic2017graph} to separately update the central node $v_i$'s hidden representation based on the mean and covariance matrices of its neighbors.

% \paragraph{Mean Aggregation}
The mean aggregation process is formulated as     $\hat{\mathbf{z}}^{mean}_i = \sigma\left( \sum_{j \in \mathcal{N}(i)} \alpha_{ij} \mathbf{W} \hat{\mathbf{\mu}}_j \right)$, where $\mathbf{W} \in \mathbb{R}^{d \times d}$ is a learnable weight matrix, $\sigma(\cdot)$ is an activation function, and $\alpha_{ij}$ is the attention coefficient between $v_i$ and $v_j$, which is computed as:
\begin{equation} 
    \alpha_{ij} = \frac{\exp\left( \text{LeakyReLU}\left(\mathbf{a}^\top [\mathbf{W}\hat{\mathbf{\mu}}_i \| \mathbf{W}\hat{\mathbf{\mu}}_j]\right)\right)}{\sum_{k \in \mathcal{N}(i)} \exp\left( \text{LeakyReLU}\left(\mathbf{a}^\top [\mathbf{W}\hat{\mathbf{\mu}}_i \| \mathbf{W}\hat{\mathbf{\mu}}_k]\right)\right)}
\end{equation}
where $\|$ is the concatenation operator and  $\mathbf{a}\in\mathbb{R}^{2d}$ is a learnable weight vector.

The covariance aggregation follows a similar process to the mean aggregation but operates on the covariance matrix of features. Given the input covariance matrix $\hat{\mathbf{\Sigma}}_i$, it is first flattened into a vector  $\hat{\mathbf{z}}^{cov}_{i} \in \mathbb{R}^{d^2}$  for processing. The aggregation process for covariance is $\hat{\mathbf{z}}^{cov}_i = \sigma\left( \sum_{j \in \mathcal{N}(i)} \alpha'_{ij} \mathbf{W}' \hat{\mathbf{z}}^{cov}_j \right)
$, where $\mathbf{W}' \in \mathbb{R}^{d \times d}$ is a learnable weight matrix, $\alpha'_{ij}$ is the attention coefficient between nodes $v_i$ and $v_j$.

The aggregated representations $\hat{\mathbf{z}}^{mean}_i$ and $\hat{\mathbf{z}}^{cov}_i$ are further transformed into the output space using MLPs:
\begin{equation} 
    \tilde{\mu}_i = \textrm{MLP}\left(\hat{\mathbf{z}}^{mean}_i\right), \quad
    \tilde{\mathbf{\Sigma}}_i = \text{Reshape}_{(d, d)}\left(\textrm{Norm}\left(\textrm{MLP}\left(\hat{\mathbf{z}}^{cov}_i\right)\right)\right)
\end{equation}
where $\tilde{\mu}_i$ is the updated mean and $\tilde{\mathbf{\Sigma}}_i$ is the updated covariance matrix for the node $v_i$. The function $\text{Reshape}(\cdot)$ reshapes the input into a $d \times d$ matrix, and $\textrm{Norm}(\cdot)$ is the normalization function.

Finally, we combine these features to update $\mathbf{z}_i\in\mathbf{Z}$ output from the joint graph convolutional encoder to obtain the updated hidden representation $\tilde{\mathbf{H}}$. Specifically, the $\tilde{\mathbf{h}}_i \in \tilde{\mathbf{H}}$ is calculated as $\tilde{\mathbf{h}}_i = (1 - \lambda_{ca})\mathbf{z}_i + \lambda_{ca}\tilde{\mu}_i \tilde{\mathbf{\Sigma}}_i$, where $\lambda_{ca}$ is a trade-off coefficient controlling the contribution of the aggregated features.

\subsection{Decoder} \label{sec:decoder}
Finally, we employ two GNN-based decoders, $\textrm{Dec}^{att}$ and $\textrm{Dec}^{str}$, to reconstruct the node attributes and adjacency matrix, respectively. These decoders take the updated hidden representations $\tilde{\mathbf{H}}$ and the graph adjacency matrix $\mathbf{A}$ as inputs. The reconstruction process is formulated as:
\begin{equation} 
    \hat{\mathbf{X}} = \textrm{Dec}^{att}(\tilde{\mathbf{H}}, \mathbf{A}), \quad \hat{\mathbf{A}} = \textrm{Dec}^{str}(\tilde{\mathbf{H}}, \mathbf{A})
\end{equation}

To evaluate the reconstruction quality, we define a loss that balances the reconstruction errors of node attributes and the adjacency matrix using a trade-off coefficient $\lambda_{\text{cs}}$:
\begin{equation} 
    \mathcal{L}_{rec} = \lambda_{cs}\mathcal{L}_{att} + (1 - \lambda_{cs})\mathcal{L}_{str}
\end{equation}
where $\mathcal{L}_{att} = \|\hat{\mathbf{X}} - \mathbf{X}\|_2$, $\mathcal{L}_{str} = \|\hat{\mathbf{A}} - \mathbf{A}\|_F$, and the coefficient $\lambda_{cs}= \frac{\textrm{std}(\mathbf{A})}{\textrm{std}(\mathbf{A}) + \textrm{std}(\mathbf{X})}$, following the method used in ADA-GAD~\cite{he2024ada}. Moreover, $\mathcal{L}_{\text{rec}}$ is also utilized as the anomaly score to quantify the anomaly degree of each node.

Finally, the total loss $\mathcal{L} = \mathcal{L}_{rec} + \mathcal{L}_{nr}$, which combines the reconstruction loss and neighbor reconstruction loss.

\section{Experiment}
\subsection{Experimental Setup}

\begin{wraptable}{r}{0.45\textwidth} 
  \centering
  \caption{Statistics of all datasets.}
  \label{tab:dataset_stats}
  \small
  \setlength{\tabcolsep}{1pt}
  \scalebox{0.9}{
  \begin{tabular}{l|rrrr}
    \toprule
    \textbf{Dataset} & \textbf{Nodes} & \textbf{Edges} & \textbf{Ano.} & \textbf{Ratio} \\ 
    \midrule
    Weibo    & 8.4k   & 408k   & 868   & 10.3\% \\
    Reddit   & 11.0k  & 168k   & 366   & 3.3\% \\
    Disney   & 124    & 335    & 6     & 4.8\% \\
    Books    & 1.4k   & 3.7k   & 28    & 2.0\% \\
    Enron    & 13.5k  & 177k   & 5     & 0.4\% \\
    Elliptic & 203k   & 234k   & 4,545 & 9.8\% \\
    DGraph   & 3.7M   & 4.3M   & 15.5k & 0.4\% \\
    \bottomrule
  \end{tabular}
  }
  \vspace{-35pt}
\end{wraptable}

\subsubsection{Datasets}
We conduct experiments on seven public real-world datasets with organically occurring anomalies, including four domains: 
social media (Weibo~\cite{zhao2020error}, Reddit~\cite{kumar2019predicting}, DGraph~\cite{huang2022dgraph}),
e-commerce (Disney~\cite{muller2013ranking}, Books~\cite{sanchez2013statistical}), 
communication (Enron~\cite{sanchez2013statistical}), and financial network (Elliptic~\cite{weber2019anti}).
The detailed statistics of all datasets are summarized in Table~\ref{tab:dataset_stats}.

\subsubsection{Baselines}
We compare the proposed NK-GAD with nine anomaly detection methods: DOMINANT~\cite{ding2019deep}, DONE~\cite{bandyopadhyay2020outlier}, 
AdONE~\cite{bandyopadhyay2020outlier}, AnomalyDAE~\cite{fan2020anomalydae}, GAAN~\cite{chen2020generative}, 
GAD-NR~\cite{roy2024gad}, ADA-GAD~\cite{he2024ada},
GADAM~\cite{chen2024boosting}, and SmoothGNN~\cite{dong2025smoothgnn}.

\subsubsection{Implementation Details}
We set the number of epochs to 30, the dropout rate to 0.3, the weight decay to 1e-5, the learning rate to 0.001 for the Reddit and Disney datasets, and 1e-4 for others. 
The embedding dimension $d$ is set to 32 for Weibo, Disney, and Books, and 16 for the other datasets. 
We repeat all experiments 10 times using 10 different seeds from 0 to 9.
In the joint graph convolutional encoder, the layer depth of the low-pass graph filter is set to 2 for Weibo, Reddit, and Books, and 1 for the others. The depth of the high-pass graph filter is set to 1 for Weibo, Reddit, and Enron datasets, and 2 for the others.
For both the contextual and structural decoders, we use GATs for Weibo and GCNs for Reddit. For others, the contextual and structural decoders are GATs and GCNs, respectively. 
To determine the appropriate values for the coefficients $\lambda_{joint}$ in the joint graph convolutional encoder and $\lambda_{ca}$ in the center aggregation, a grid search is conducted for each dataset. The values of $\lambda_{joint}$ and $\lambda_{ca}$ are searched over the range from 0.1 to 0.9 in steps of 0.1. 
The analysis of these hyperparameters is provided in Sec.~\ref{sec:hyperparameter}.

For large-scale datasets such as Elliptic and DGraph, both the proposed NK-GAD and reconstruction-based baselines (e.g., ADA-GAD and GAAN) adopt a mini-batch scheme~\cite{liu2022bond} to reduce memory consumption. The complexity analysis of NK-GAD is provided in Sec.~\ref{sec:complexity}.

We use AUC (Area Under the Curve) as the metric to evaluate the performance of all methods~\cite{liu2022bond,he2024ada,roy2024gad}. 

\begin{table*}[!ht]
\caption{AUC (\%) results (mean$\pm$std) across seven datasets. The best results are highlighted in bold, while sub-optimal results are underlined. OOM denotes out of memory with regard to CPU~\cite{liu2022bond}.}
\centering
\setlength{\tabcolsep}{1pt}
\scalebox{0.8}{
\begin{tabular}{c|ccccccc}
    \midrule[1.5pt]
        Method & Weibo & Reddit & Disney & Books & Enron & Elliptic & DGraph \\ 
        \midrule[1.5pt]
        DOMINANT  & 89.45$\pm$1.76 & 56.10$\pm$0.05 & 49.36$\pm$4.32 & 57.30$\pm$4.53 & 52.95$\pm$3.88 & OOM\_C & OOM\_C\\
        DONE & 84.20$\pm$1.67 & 54.90$\pm$1.28 & 42.32$\pm$5.27 & 46.80$\pm$6.35 & 48.71$\pm$6.43 & OOM\_C &  OOM\_C \\
        AdONE & 83.50$\pm$3.05 & 53.89$\pm$2.51 & 48.22$\pm$2.64 & 54.36$\pm$1.07 & 51.79$\pm$1.69 & OOM\_C & OOM\_C \\
        AnomalyDAE & 89.88$\pm$1.32 & 55.31$\pm$1.65 & 48.45$\pm$2.46 & 58.87$\pm$1.42 & 48.54$\pm$2.89 & OOM\_C & OOM\_C \\
        GAAN & 89.82$\pm$0.03 & 55.34$\pm$0.61 & 48.02$\pm$0.00 & 58.09$\pm$2.57 & 56.77$\pm$5.00 & OOM\_C & OOM\_C\\
        ADA-GAD  & \underline{92.44$\pm$0.46} & {56.89$\pm$0.01} & 66.37$\pm$4.83 & \underline{64.43$\pm$3.74} & 69.15$\pm$4.26 & \underline{49.85$\pm$2.75} & 48.99$\pm$1.16 \\
        GAD-NR & 73.81$\pm$2.01 & 56.77$\pm$2.92 & {69.83$\pm$7.48} & 58.79$\pm$6.87 & \underline{73.44$\pm$9.43} & 44.63$\pm$7.33 & 47.63$\pm$1.21 \\
        GADAM & 25.54$\pm$0.85 & \textbf{57.75$\pm$0.64} & \underline{70.72$\pm$1.44} & 59.13$\pm$5.40 & 36.08$\pm$1.50 & 46.89$\pm$1.08 & 45.22$\pm$0.82\\
        SmoothGNN & 56.97$\pm$8.72 & 53.76$\pm$3.39 & 54.01$\pm$10.92 & 48.16$\pm$5.53 & 51.48$\pm$5.47 & 48.00$\pm$5.02 & \underline{50.44$\pm$5.20} \\
        NK-GAD & \textbf{93.70$\pm$0.87} & \underline{57.27$\pm$0.07} & \textbf{77.26$\pm$2.25} & \textbf{65.60$\pm$0.72} & \textbf{80.82$\pm$3.10} & \textbf{52.17$\pm$3.41} & \textbf{55.30±0.16}\\
        \midrule[1.5pt]
    \end{tabular}
}
\label{table:main}
\end{table*}

\begin{table}[!ht]
\centering
\caption{AUC (\%) results of NK-GAD and its three variants. Best results are in bold; sub-optimal ones are underlined.}
\setlength{\tabcolsep}{3pt}
\scalebox{0.9}{
\begin{tabular}{c|ccccccc}
    \midrule[1.5pt]
        Method & Weibo & Reddit & Disney & Books & Enron & Elliptic & DGraph \\ 
        \midrule[1.5pt]
        NK-GAD$^\dag$ & 89.57 & 55.32 & 68.64 &  60.61 & 63.16 & 49.73 & 48.44 \\
        NK-GAD$^\ddag$ & 91.21 & 56.21 & 76.21 & 64.74 & 79.88 & 51.26 & 49.32 \\
        NK-GAD$^\S$ & \underline{92.71} & \underline{56.92} & \underline{76.40} & \underline{65.18} & \underline{80.21} & \underline{51.49} & \underline{51.44} \\
        NK-GAD & \textbf{93.70} & \textbf{57.27} & \textbf{77.26} & \textbf{65.60} & \textbf{80.82} & \textbf{52.17} & \textbf{55.30} \\
        \midrule[1.5pt]
    \end{tabular}
}
\label{table:ablation}
\end{table}

\subsection{Performance Analysis}

As shown in Tab.~\ref{table:main}, the proposed NK-GAD consistently achieves superior performance across six of the seven datasets, yielding an average 3.29\% AUC improvement over the best baselines. Specifically, NK-GAD attains the highest AUC on six datasets, especially the largest datasets, Elliptic and DGraph. These consistent gains demonstrate that NK-GAD effectively enhances its robustness across diverse graph domains.
The only marginal decline appears on the Reddit dataset, where NK-GAD slightly underperforms GADAM, due to indistinct high-frequency structural patterns that weaken spectral modeling advantages. Nevertheless, NK-GAD surpasses all other baselines and achieves the highest overall average AUC (68.16\% vs.\ 62.73\% for ADA-GAD), confirming its strong generalization capability in graph anomaly detection tasks.

\subsection{Ablation Study}
To evaluate the contributions of each module in NK-GAD, we design three variants: NK-GAD$^\dag$ replaces the joint graph convolutional encoder with a plain GCN-based encoder and removes both the neighbor reconstruction and center aggregation modules; NK-GAD$^\ddag$ retains the joint graph convolutional encoder but excludes the neighbor reconstruction and center aggregation modules; NK-GAD$^\S$ removes only the center aggregation module, keeping the joint graph convolutional encoder and neighbor reconstruction module retained.
Tab.~\ref{table:ablation} highlights the role of each component. 
The performance gap between NK-GAD$^\ddag$ and NK-GAD$^\dag$ demonstrates that the joint graph convolutional encoder is effective on datasets with strong high-frequency signals, such as Disney and Enron. Incorporating neighbor reconstruction and center aggregation further enhances detection on graphs with a large amount of edges, like Weibo and DGraph, where numerous neighbors introduce redundant patterns that require refinement.

\subsection{Hyperparamter Analysis} \label{sec:hyperparameter}
\begin{figure}[!ht]
	\centering
	\subfloat[Joint Graph Conv Encoder]{
		\includegraphics[width=0.4\textwidth]{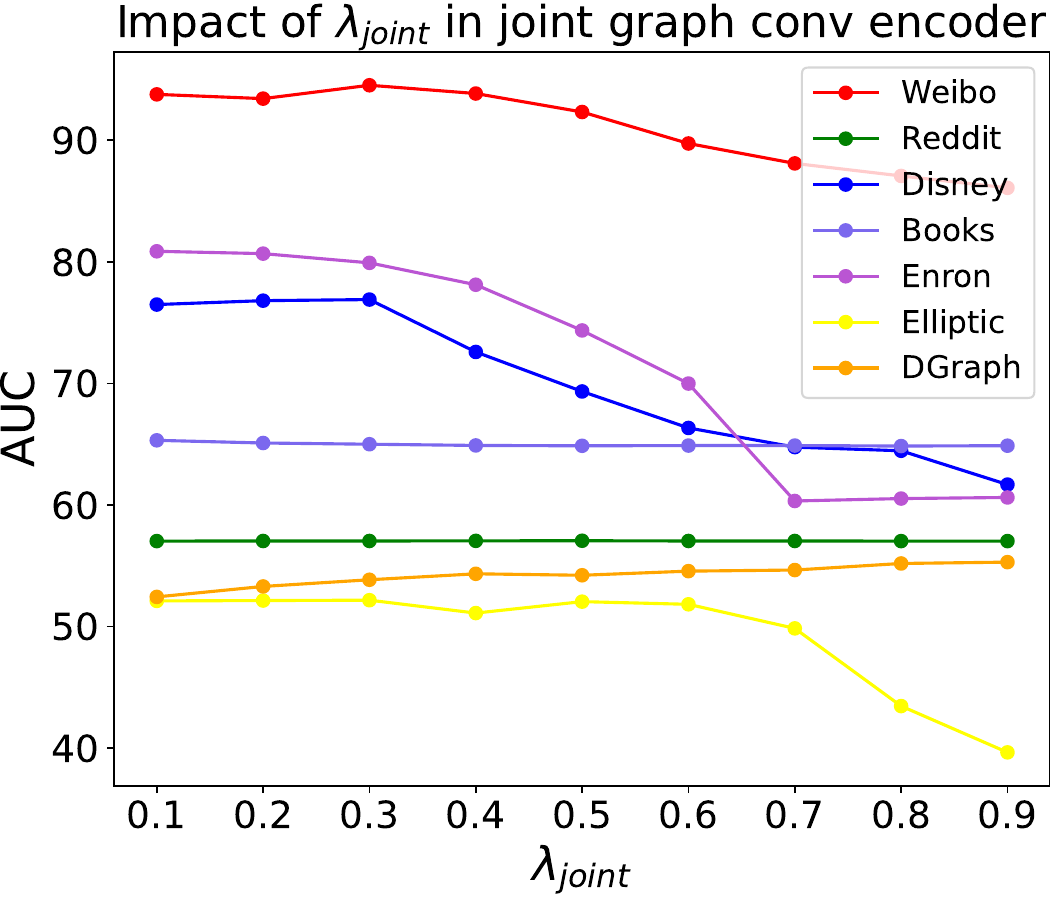} 
		\label{fig:high_coe}} 
	\subfloat[Center Aggregation]{
		\includegraphics[width=0.4\textwidth]{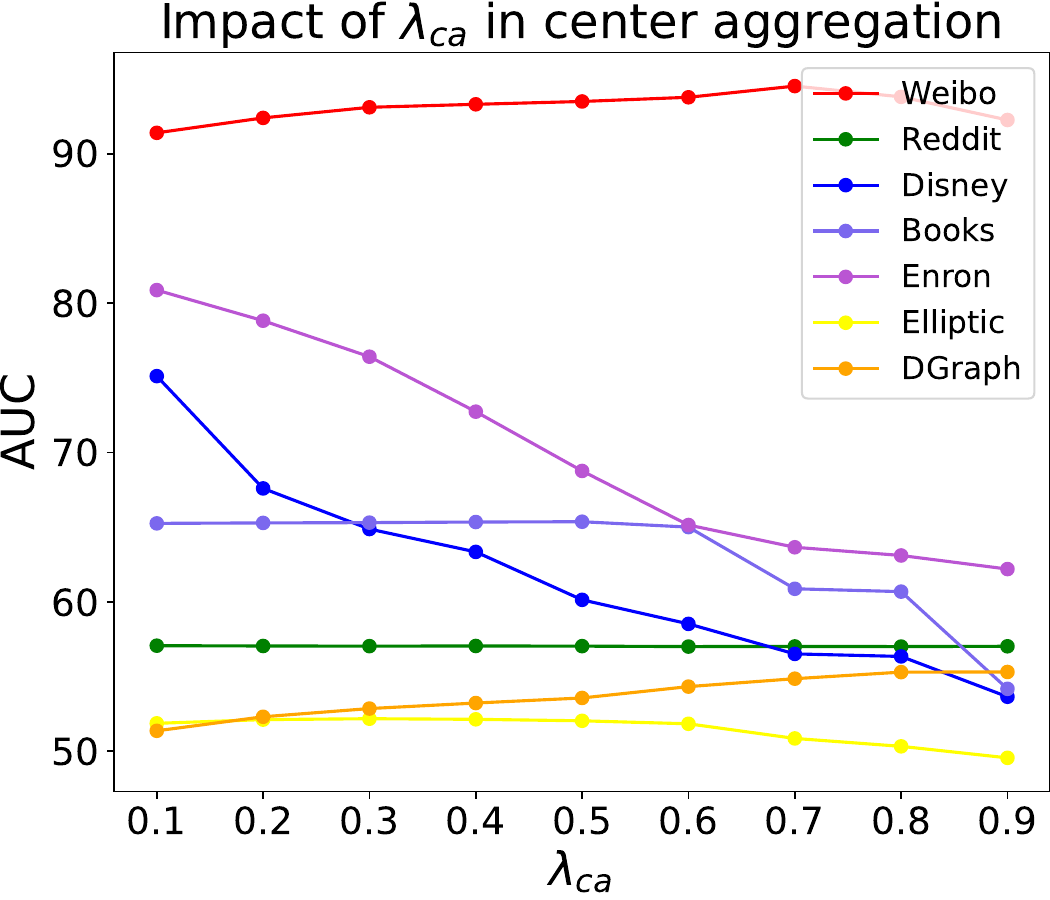}
		\label{fig:center_rec_coe}}
	\caption{Impacts of $\lambda_{joint}$ in the joint graph convolutional encoder and $\lambda_{ca}$ in the center aggregation module across datasets.}
	\label{fig:hyperparameter}
\end{figure}

We conduct extensive experiments to analyze the impact of the hyperparameters $\lambda_{joint}$ and $\lambda_{ca}$ across all datasets, as shown in Figure~\ref{fig:hyperparameter}.
For $\lambda_{joint}$, NK-GAD shows a performance drop on datasets such as Disney and Enron as the value increases, indicating high sensitivity to high-frequency components. In contrast, performance on Reddit and Books remains stable, suggesting limited dependence on this parameter. These results highlight the importance of balancing high- and low-frequency information in a dataset-specific manner.
For $\lambda_{ca}$, performance on Disney and Books declines notably beyond intermediate values, reflecting diminishing gains from overemphasizing updated central node features. Conversely, larger graphs like Weibo and DGraph benefit from higher $\lambda_{ca}$, where numerous neighbors introduce redundant patterns that require refinement.

\subsection{Computational Complexity Analysis}  \label{sec:complexity}

We further analyze the computational complexity of the proposed NK-GAD. 
The joint graph convolutional encoder processes low-pass and high-pass filters over $L$ and $L'$ layers, respectively. Each layer involves sparse graph convolution with complexity $O(|\mathcal{E}|d)$ and linear transformations with complexity $O(|\mathcal{V}|d^2)$, leading to a total encoder complexity of $O\left((L + L') \cdot (|\mathcal{E}|d + |\mathcal{V}|d^2)\right)$, where $|\mathcal{E}|$, $|\mathcal{V}|$, and $d$ denote the number of edges, nodes, and feature dimensions. The neighbor reconstruction module computes self-feature reconstruction ($O(|\mathcal{V}|d^2)$) and neighbor statistics (mean: $O(|\mathcal{E}|d)$, variance/covariance: $O(|\mathcal{E}|d^2)$), resulting in $O\left(|\mathcal{E}|d^2 + |\mathcal{V}|d^2\right)$. The center aggregation module employs graph attention networks to update node features, requiring $O(|\mathcal{E}|d)$ for attention scores and $O(|\mathcal{V}|d^2)$ for feature updates, with total complexity $O\left(2 \cdot (|\mathcal{E}|d + |\mathcal{V}|d^2)\right)$. For decoders, the attribute decoder incurs $O(|\mathcal{E}|d + |\mathcal{V}|d^2)$, while the structure decoder reconstructs the adjacency matrix with $O(|\mathcal{V}|^2d)$. The overall complexity is dominated by $O\left(|\mathcal{E}| + |\mathcal{V}|^2\right)$. The complexity is equal to the reconstruction-based methods in baselines like ADA-GAD and GAD-NR.

To adapt to large graphs such as Elliptic and DGraph, NK-GAD is designed to be efficiently deployed on local GPUs through a mini-batch strategy~\cite{liu2022bond}. 
By sampling node subsets with their $k$-hop neighborhoods, the model reduces memory complexity from quadratic to near-linear in $|\mathcal{V}|$, 
enabling scalable learning on million-node graphs. 
Under this setting, NK-GAD remains superior to full-graph-processed baselines, showing that neighborhood-knowledge aggregation is both effective and deployable on large-scale attribute-heterophily graphs.

\section{Conclusion}
This paper reveals two phenomena: 1) attribute similarities between connected nodes show nearly identical distributions across different connected node pair types, and 
2) anomalies cause consistent variation trends between the graph with and without anomalous edges in the both low- and high-frequency components of the spectral energy distributions, while the mid-part exhibits more erratic variations. 
Building on these insights, we propose NK-GAD that leverages neighbor knowledge for graph anomaly detection. 
A joint graph convolutional encoder captures high- and low-frequency patterns, while a neighbor reconstruction module learns normal neighbor distributions to refine detection, complemented by a center aggregation module for feature updating.  Extensive experiments on seven real-world datasets demonstrate the effectiveness of NK-GAD.

\subsubsection{Acknowledgments.}
This work was supported in part by the National Natural Science Foundation of China under Grant 62572346.

\bibliographystyle{splncs04}
\bibliography{reference}
\end{document}